\documentclass[runningheads]{llncs}
\usepackage{graphicx}
\usepackage{comment}
\usepackage{amsmath,amssymb} 
\usepackage{color}

\usepackage{subfigure}
\usepackage{enumitem}
\usepackage{booktabs}
\usepackage{multirow}
\usepackage{seqsplit}

\newcommand{\modelname}[0]{STAR}

\begin{document}
\pagestyle{headings}
\mainmatter
\def\ECCVSubNumber{7082}  

\title{Spatio-Temporal Action Detection with Multi-Object Interaction}

\titlerunning{Spatio-temporal Action Detection with Multi-Object Interaction}

\author{Huijuan Xu$^1$ \hspace{1.5mm} 
Lizhi Yang$^1$  \hspace{1.5mm}
Stan Sclaroff$^2$ \hspace{1.5mm}
Kate Saenko$^2$  \hspace{1.5mm}
Trevor Darrell$^1$
}

\authorrunning{Huijuan Xu et al.} 

\institute{$^1$University of California, Berkeley  \qquad  \qquad
$^2$Boston University \\
\email{
$^1$\{huijuan, lzyang, trevor\}@eecs.berkeley.edu,
$^2$\{sclaroff, saenko\}@bu.edu}
}

\maketitle

\begin{abstract}
Spatio-temporal action detection in videos requires localizing the action both spatially and temporally in the form of an ``action tube.''
Nowadays, most spatio-temporal action detection datasets (e.g. UCF101-24, AVA, DALY) are annotated with action tubes that contain a single person performing the action, thus the predominant action detection models simply employ a person detection and tracking pipeline for localization.
However, when the action is defined as an interaction between multiple objects, such methods may fail since each bounding box in the action tube contains multiple objects instead of one person.
In this paper, we study the spatio-temporal action detection problem with multi-object interaction.
We introduce a new dataset that is annotated with action tubes containing multi-object interactions. 
Moreover, we propose an end-to-end spatio-temporal action detection model that performs both spatial and temporal regression simultaneously. Our spatial regression may enclose multiple objects participating in the action.
During test time, we simply connect the regressed bounding boxes within the predicted temporal duration using a simple heuristic.
We report the baseline results of our proposed model on this new dataset,
and also show competitive results on the standard benchmark UCF101-24 using only RGB input.
\end{abstract}

\vspace{-10pt}
\section{Introduction}
\begin{figure}[!t]
\centering
\includegraphics[width=0.8\linewidth]{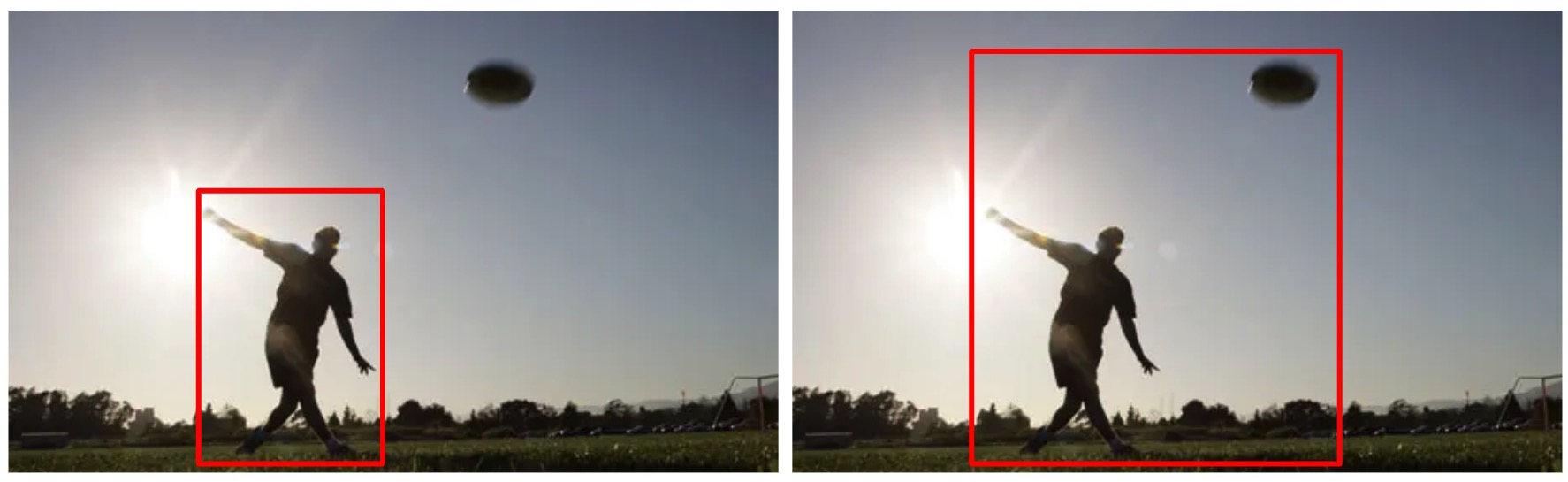}
\vspace{-0.1in}
\caption{Toy example for annotating the action ``a person throws the frisbee". Existing datasets annotate the action around the person subject (left), while we take the action definition of subject and object interaction and annotate the action enclosing the involved subjects and objects (right).}
\vspace{-0.1in}
\label{fig:toy_example}
\end{figure}

\begin{figure*}[!t]
\centering
\subfigure[Moving something and something away from each other]{
\includegraphics[width=0.85\linewidth]{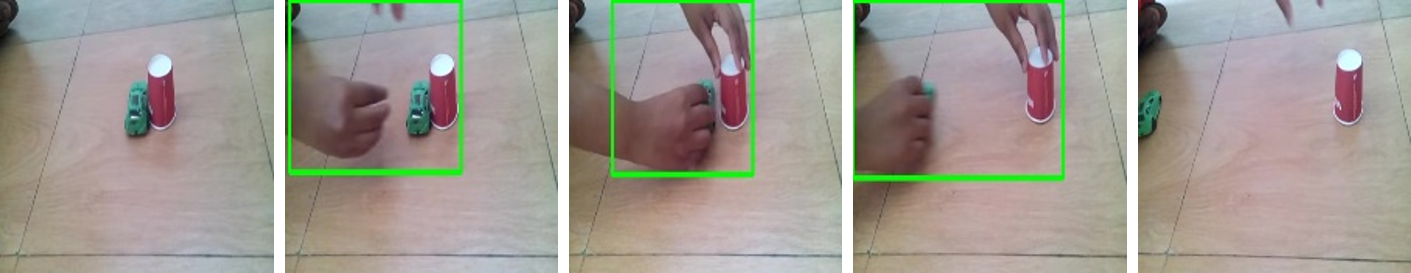}
\label{fig:smth-move-away}
}
\subfigure[Moving something and something closer to each other]{
\hbox{\hspace{0.14cm}\includegraphics[width=0.85\linewidth]{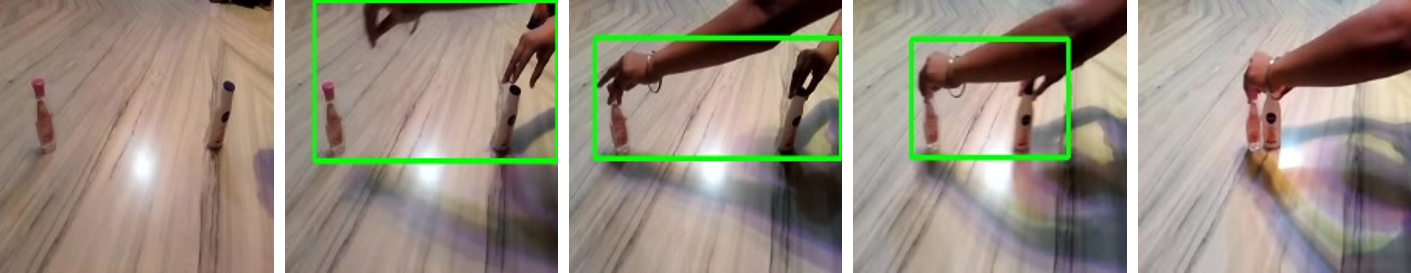}}
\label{fig:smth-move-close}
}
\vspace{-0.1in}
\caption{Two example videos from our something-STAR spatio-temporal action detection dataset with the action bounding box containing multi-object interaction.}
\vspace{-0.1in}
\label{fig:distinguish-multi-object}
\end{figure*}

Current methods for spatio-temporal action detection mostly focus on human-centric actions.
The bounding boxes in corresponding datasets (e.g. UCF101-24~\cite{soomro2012ucf101}, AVA~\cite{gu2018ava}, DALY dataset~\cite{weinzaepfel2016human}) are annotated around the human subject.
This raises the question of how actions should be defined.
Should they be defined as the movement of the subject (e.g. person) or as the interaction of involved subjects and objects?
Take the action ``a person throws the frisbee" as a toy example, shown in Figure~\ref{fig:toy_example}. Should the annotated action bounding box be focused on the person throwing the frisbee or includes both the person and the frisbee? The answer probably is the latter.
In this paper we take the definition of action as a subject/object interaction, and propose a spatio-temporal action detection dataset called something-STAR dataset based on the something-something video classification dataset~\cite{goyal2017something}.
Our bounding box annotation for the action tube includes the action subject (hand) and the involved objects.
Figure~\ref{fig:distinguish-multi-object} shows two example videos from our annotated dataset with the action tube annotation containing multi-object interaction, namely the actions ``Moving something and something away from each other" and ``Moving something and something closer to each other".

Existing methods using the object detection and linking pipeline~\cite{gkioxari2015finding,weinzaepfel2015learning} are not well suited to this type of action detection with multi-object interaction, since they only rely on the object detection models to realize spatial localization. However, reliable detectors may not be available for all objects.
In this paper, we propose a simple, elegant and effective multi-object action detection model with simultaneous spatial and temporal regression, called the \textit{Spatio-TemporAl Regression (STAR)} model.
Our model regresses the spatial action bounding box containing multi-object interaction without replying on external object detectors.
It also contains a temporal localization branch to regress the temporal duration in parallel with the spatial regression during training time.
At test time, previous action detection models link the detected bounding boxes and realize temporal localization using heuristic optimization methods~\cite{yu2015fast,saha2016deep,singh2017online,zhu2017tornado} or a sliding window approach~\cite{weinzaepfel2015learning}. Our STAR model takes the opposite top-down approach by first selecting a generated temporal segment proposal and then connecting the regressed spatial action boxes within the temporal duration using a simple intersection over union heuristic without extra optimization.

Besides using the linking heuristic~\cite{yu2015fast,saha2016deep,singh2017online,zhu2017tornado} or the sliding window approach~\cite{weinzaepfel2015learning}, recent methods try to improve temporal localization by modeling the transition state~\cite{song2019tacnet} or by gradually extending the temporal context~\cite{yang2019step}.
In contrast to these expensive and complicated temporal solutions, we propose direct temporal regression and integrate it into the training process to guide the action detection with multi-object interaction, which is very simple and effective.
Another drawback of current spatio-temporal action detection methods is that their classification backbones are two-stream architectures using both RGB and flow inputs to provide motion context.
Notably, the flow computation is expensive, thus several recent papers~\cite{zhao2019dance,su2019improving,pramono2019hierarchical} try to merge the flow stream with the RGB stream to reduce computation cost.
In this paper, we only work on RGB input with three-dimensional convolution to aggregate motion context.
We use the shared feature maps from the three-dimensional convolution output to regress 
action tubes containing multiple object interaction,  significantly improving detection speed.
Last but not least, our model is end-to-end trainable and thus straightforward to deploy.

To summarize, our contributions are:
\vspace{-7pt}
\begin{itemize} [leftmargin=*]
    \itemsep0em
    \item We introduce a simple and effective spatio-temporal action detection model called STAR with both spatial and temporal regression; our model achieves efficiency by simply connecting the boxes within the regressed temporal duration during test time, without heavy post-processing;
    \item We propose a new large-scale spatio-temporal action detection dataset with multi-object interaction in the annotated action tube;
    \item We validate the effectiveness of our proposed STAR model on our new dataset and on the UCF101-24 benchmark, achieving competitive results.
\end{itemize}

\vspace{-10pt}
\section{Related Work}
\label{Related}

Spatio-temporal action detection is a challenging problem as it requires localizing the action both spatially and temporally in the form of action tube.
Early approaches define action filters~\cite{ke2005efficient,rodriguez2008action} or part-based templates~\cite{ke2007event}, and then detect actions through matching in sliding subvolume way, which lacks generalization ability.
Another category of models group segmented supervoxels~\cite{soomro2015action,jain2014action} or cluster dense trajectories~\cite{chen2015action,van2015apt} to detect actions showing better robustness compared to previous template based approaches.
Also, multiple works~\cite{wang2014video,ma2013action,lan2011discriminative} model the part relationship through structure SVM.
Noticeably, early datasets on this task contain limited amount of classes, e.g. six classes for KTH dataset~\cite{schuldt2004recognizing} and three classes for MSR dataset~\cite{cao2010cross}.
Some other action detection datasets (e.g. UCF-Sports~\cite{rodriguez2008action}, J-HMDB~\cite{jhuang2013towards} and AVA~\cite{gu2018ava}) focus on spatial localization with trimmed videos. 
Only UCF101-24~\cite{soomro2012ucf101} and DALY~\cite{weinzaepfel2016human} have both spatial and temporal annotation, but DALY's spatial annotation is sparse annotation on selected frames within the action duration.

Moreover, almost all of these action datasets focus on human centric actions, which delivers another category of action detection models based on object detection. Specific models include generalizing Deformable part model (DPM) for 2D image object detection to 3D volumes~\cite{tian2013spatiotemporal}.
Among them the most dominant approaches take the two-stage pipeline with object detection and linking optimization following the paper~\cite{gkioxari2015finding}.
While the two-stage pipeline may be suitable for the human centric action detection, in the case of action detection with multiple object interaction this type of methods may fail since the single object bounding box is not equal to the bounding box enclosing the whole action.
In this paper, we explore the spatio-temporal action detection with multi-object interaction from both dataset and model aspect.

Temporal localization is one important factor in spatio-temporal action detection. \cite{gkioxari2015finding} links the detection boxes per frame into action tube on trimmed videos without paying attention to temporal localization.
Other models either take one extra temporal sliding stage over the action tubes~\cite{weinzaepfel2015learning} or design dynamic optimization in the box linking process~\cite{yu2015fast,saha2016deep,singh2017online,zhu2017tornado} to realize temporal localization in spatio-temporal action detection task. Recently, new approaches~\cite{song2019tacnet,yang2019step} are proposed with specific designs to improve the temporal localization part in spatio-temporal action detection.
The TACNet~\cite{song2019tacnet} includes two designed components to facilitate temporal localization, namely the temporal context detector encoding the long term context information and the transition aware classifier distinguishing transitional states.
The Spatio-Temporal Progressive (STEP) action detector~\cite{yang2019step} progressively processes longer sequences and adaptively extends proposals to follow the action movement. 
In the task of temporal action detection which only requires to temporally localize the action start time and end time, direct temporal coordinate regression is already shown to be a very effective temporal localization mechanism~\cite{xu2017r,chao2018rethinking}.
However, no existing spatio-temporal action detection models take direct temporal regression as their temporal localization solution. Though some spatio-temporal action detection models~\cite{saha2017amtnet,kalogeiton2017action,hou2017tube} extend single bounding box regression to cuboid regression consisting of multiple bounding box regression simultaneously, they still rely on the short tubelet linking optimization to realize temporal localization.
In this paper, we are the first to integrate direct temporal regression into the
spatio-temporal model training stage and show good performance for this simple temporal localization solution.

Now most spatio-temporal action detection models are built on top of the two-stream video classification backbone~\cite{simonyan2014two} with both RGB and flow inputs to incorporate motion context.
Some recent models facilitate the communication between the two streams
through progressive cross-stream message passing~\cite{su2019improving} or hierarchical self-attention~\cite{pramono2019hierarchical} to learn better feature representation for better action detection.
\cite{zhao2019dance} proposes to embed RGB and optical-flow into a single two-in-one stream network to reduce heavy computation cost. A motion modulation layer is designed to generate transformation parameters from flow images to modulate low-level RGB features.
In our proposed model, we directly take three-dimensional convolution on the RGB input~\cite{tran2015learning} as feature extraction backbone, and on the shared convolution feature output, two independent branches are designed for temporal segment regression and bounding box spatial regression in the action tube.
Moreover, our spatial bounding boxes can be regressed according to the spatial annotation of action tube containing multiple object interaction and are learned with the extra supervision of action classification to capture the action motion pattern.
Therefore, our model fits both human centric action detection and action detection with multi-object interaction as we show on the UCF101-24 benchmark and our proposed action detection dataset with multi-object interaction.

\section{Spatio-TemporAl Regression (STAR) Model}
\label{sec:approach}

We first formulate the problem definition for spatio-temporal action detection in videos. Given a sequence of video frames ${I_t},t=1..T$, the spatio-temporal action detection task requires to output a series of bounding boxes ${R_t},t=ts..te$ between a start time $ts$ and an end time $te$ enclosing the action both spatially and temporally (called action tube), and classify the action tube into one action class $c \in C$ where $C$ is the set of action classes. ${R_t}$ represents the four coordinates $\{x1, y1, x2, y2\}$ of the action bounding box at frame $I_t$.

\begin{figure*}[ht]
\begin{center}
\includegraphics[width=1\linewidth]{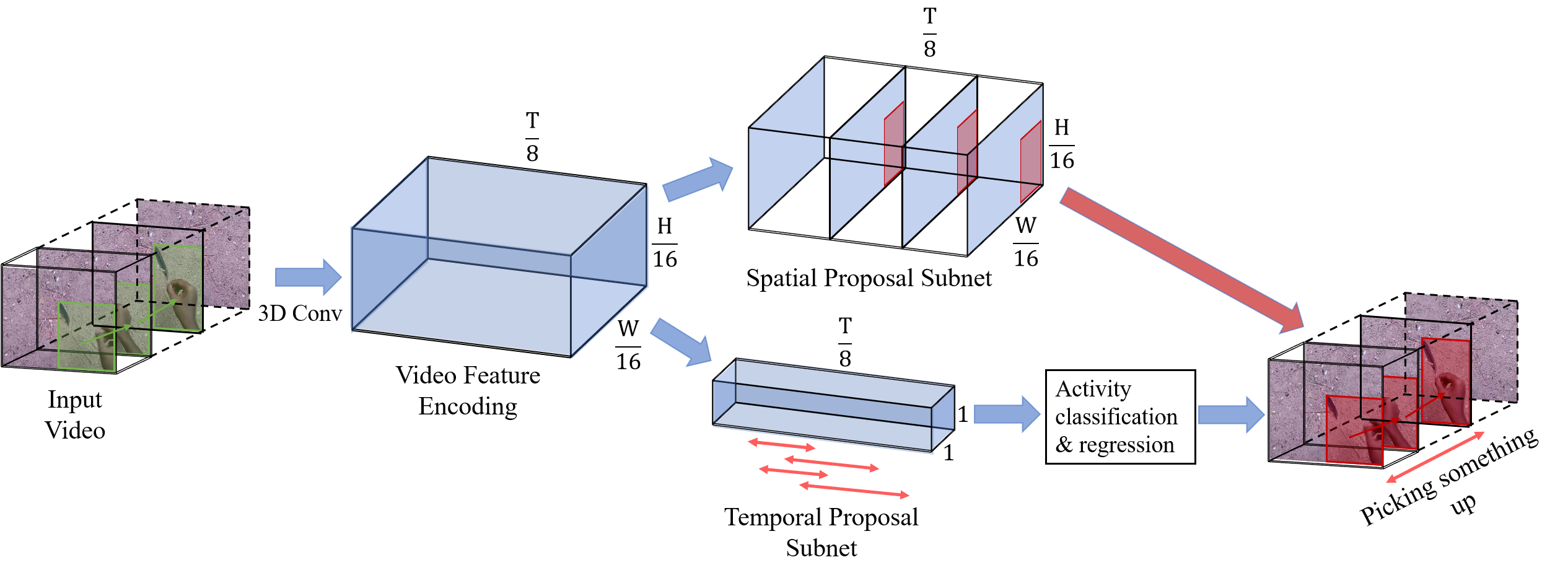}
\end{center}
\vspace{-0.1in}
\caption{Our STAR model architecture.
The 3D ConvNet takes raw video frames as input and computes convolutional features which are shared between the saptial proposal subnet and temporal proposal subnet.
Our model conducts spatial and temporal regression as well as action classification simultaneously during training, and during testing the final action detection is achieved by connecting the predicted spatial bounding boxes within the predicted temporal segment using simple IoU heuristic (shown in red color). }
\vspace{-0.1in}
\label{fig:architecture}
\end{figure*}

To tackle this challenging problem in untrimmed video streams, we propose a novel \textit{Spatio-TemporAl Regression} model called \textit{STAR} in this paper.
We are the first spatio-temporal action detection model with temporal coordinate regression as the final temporal localization policy and additionally we incorporate a spatial regression branch for simultaneous action bounding box regression of multiple objects, which enables our STAR model to detect action tube with multi-object interaction and achieves good performance.
The network, illustrated in Figure~\ref{fig:architecture}, consists of a shared 3D ConvNet feature extractor~\cite{tran2015learning}, and on top of which, a temporal localization and activity classification branch and a spatial localization branch are proposed for action tube detection and classification.
During training time, the temporal localization and spatial localization of action tubes are optimized in two independent branches, and during testing time, we ensemble the predicted temporal segments and spatial action bounding boxes into action tubes.
Our model shares the 3D video feature encoding between the temporal branch and the spatial branch, and thus enables efficient computation and end-to-end training with easy deployment.
Next, we describe our STAR model in Section~\ref{sec:model}, and Section~\ref{sec:implementation_detail} details the optimization strategy during training and the inference process during testing.


\subsection{STAR Model}
\label{sec:model}
The design of STAR model consists of simultaneous spatial action bounding box regression and temporal segment regression.
We borrow good practice from the previous temporal activity detection model (R-C3D~\cite{xu2017r}) and object detection model (Faster R-CNN~\cite{ren2015faster}) as our design choice.
We briefly recap some components from R-C3D~\cite{xu2017r} and Faster R-CNN~\cite{ren2015faster} when describing our STAR model. For more details, please refer to the original papers~\cite{xu2017r,ren2015faster}.

First, regarding the video feature encoding, our model takes a sequence of RGB video frames ${I_t},t=1..T$ with dimension $\mathbb{R}^{3\times T \times H\times W}$. Then, a 3D ConvNet~\cite{tran2015learning} is used to extract rich spatio-temporal feature hierarchies with the output feature map $C_{conv5b}\in \mathbb{R}^{D\times \frac{T}{8} \times \frac{H}{16}\times \frac{W}{16}}$, where $D$ is the channel dimension.
We use $C_{conv5b}$ feature activations as the shared input to the temporal localization and classification branch and the spatial action localization branch.

\subsubsection{Spatial localization branch}
Our spatial localization branch conducts the action bounding box regression at each frame and outputs binary actionness score for each regressed box.
We take the region proposal network of faster R-CNN~\cite{ren2015faster} as our spatial localization branch.
The number of encoding feature maps $\frac{T}{8}$ is treated as batch size dimension during the training of spatial localization, and each feature map conducts spatial localization independently. 
The ground truth action tube bounding boxes ${R_t},t=ts..te$ are mapped to one of the $\frac{T}{8}$ feature maps as spatial supervision according to nearest neighbour policy.
For a certain feature map, if no ground truth action tube is temporally across that feature map, it will not contribute to the spatial localization training.

\subsubsection{Temporal localization and classification branch}
Our temporal localization branch conducts the temporal segment regression for the action tube and classifies the action temporal segment into specific action classes.
We take the same temporal regression and classification design of R-C3D~\cite{xu2017r} in this branch of our STAR model.
Namely, the video encoding feature map is spatially pooled and on top of which, class-agnostic temporal proposals are predicted with respect to anchor segments, and high quality temporal proposals are refined and classified into specific action classes.

Notably, our spatial regression branch and temporal regression branch are two independent branches operating on the shared video feature encoding feature maps.
Our spatial regression branch only outputs class-agnostic actionness score, while the final action classification is achieved in the temporal branch, and the features used for final action classification come from 3D segment of interest (SoI) pooling for selected temporal proposals.
Thus, we use whole scene features for the action classification in our STAR model and through experiments we show good results in this first attempt to design one spatio-temporal action detection model with both spatial regression and temporal regression.
Though more complicated design for constructing intermediate action tubes during training and using the tube features instead of whole scene features for final action classification could be further explored, we leave it for future work.
In this paper, with the simple parallel spatial and temporal regression design and using whole scene features within the temporal proposals for final action classification, we already show good performance compared to previous two-stage pipelines for action detection.

\subsection{Implementation Details}
\label{sec:implementation_detail}
\subsubsection{Training Optimization}
We inherit the network architecture from existing R-C3D~\cite{xu2017r} model and the region proposal network of faster R-CNN~\cite{ren2015faster}.
We train the network by optimizing both the classification and regression tasks jointly in the two branches of our STAR model.
The softmax/binary classification loss function is used for classification, and smooth L1 loss function~\cite{girshick2015fast} is used for regression.
We also follow the standard practice in previous works~\cite{ren2015faster,xu2017r} to assign positive/negative labels to the anchors, and set positive/negative ratio in sampled batches.
We train our model on a single GPU for 250k iterations with a learning rate of 0.001. We use SGD solver with a weight decay of 0.0005 and a momentum of 0.9. The spatial anchors span 4 scales and 5 aspect ratios. The temporal segment proposal anchors are set according to specific dataset.

\subsubsection{Inference}
At test time, the temporal proposal number is 300 and the spatial proposal number is 50. We run non-maximum suppression(NMS)~\cite{girshick2015deformable} on both the temporal proposals and the spatial proposals with threshold 0.4 and 0.2.
During testing, the action tube detection in our STAR model is achieved by assembling the regressed temporal segment proposals and spatial bounding boxes in each frame.
We first take one temporal segment proposal
and then we connect the bounding boxes within the temporal segment to construct action tube starting from the first frame of the temporal segment and using a simple heuristic of maximum bounding box IoU overlap. 
The action tube score is defined by the sum of the activity classification score and the averaged bounding box proposal scores within the predicted temporal segment.

\section{Action Detection Dataset with Multi-Object Interaction}
\label{sec:dataset}
In this paper, we explore the spatio-temporal action detection in videos with multi-object interaction. To tackle this problem, we introduce one dataset called something-STAR.
The videos used in our something-STAR dataset comes from the ``seqsplit{something-something}" dataset~\cite{goyal2017something}.
Notably, labels in the original ``something-something" dataset are template descriptions with something as placeholders for objects, and crowd-workers choose the objects in the template to act the actions.
Therefore, videos in "something-something" dataset contains rich interaction with various objects and are good video source for our spatio-temporal action detection dataset with multi-object interaction.
Moreover, due to the template property of action labels without specific object name, it will encourage the action detection model to learn the motion pattern form object iteration instead of paying attention to the specific objects involved.

\begin{figure*}[!t]
\centering
\subfigure[AVA: Dance]{
\includegraphics[width=0.85\linewidth]{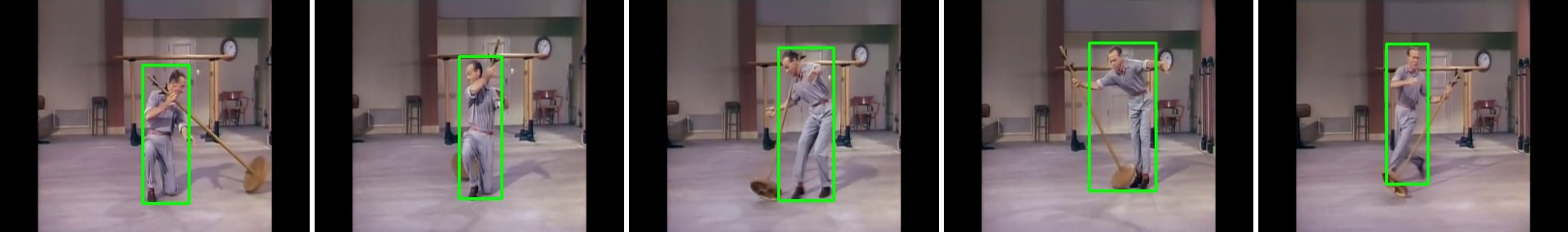}
\label{fig:ava-dance}
}
\subfigure[DALY: Drinking]{
\includegraphics[width=0.85\linewidth]{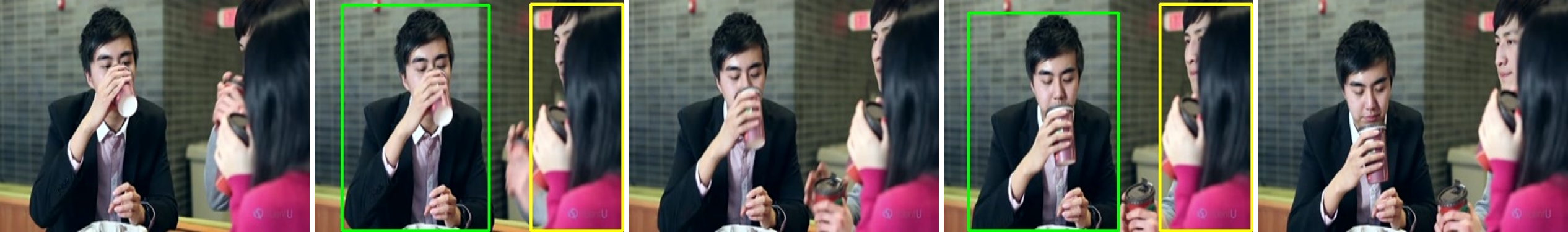}
\label{fig:DALY-drinking}
}
\subfigure[UCF101-24: Trampoline Jumping]{
\hbox{\hspace{0.01em}\includegraphics[width=0.85\linewidth]{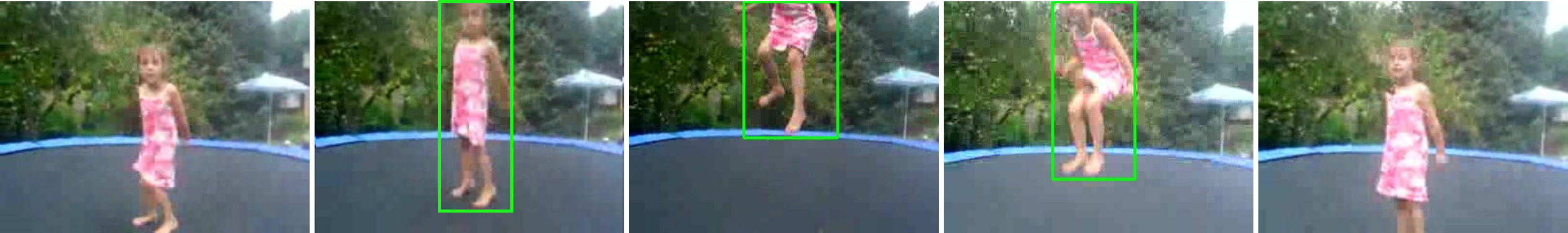}}
\label{fig:ucf-24-trampolineJumping}
}
\subfigure[Our something-STAR: Putting something next to something]{
\hbox{\hspace{0.1cm}\includegraphics[width=10.4cm,height=0.125\linewidth]{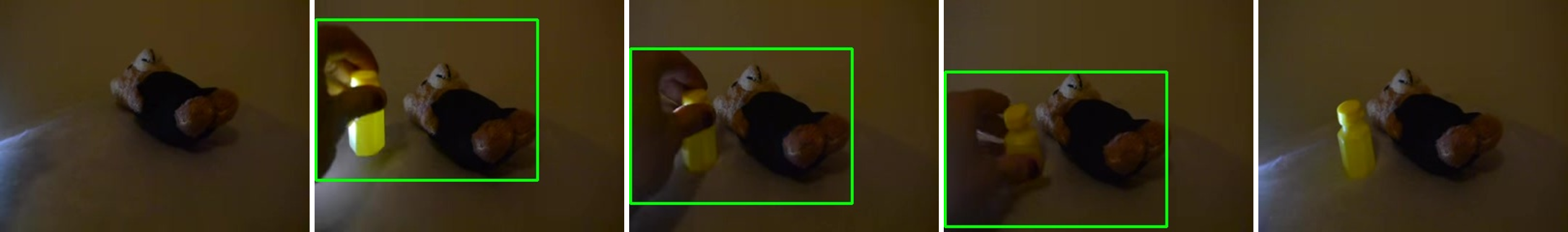}}
\label{fig:something-STAR-64}
}
\vspace{-0.1in}
\caption{Visualization of action tube annotation from existing spatio-temporal action detection datasets (e.g. AVA, DALY and UCF101-24) and our something-STAR dataset.}
\vspace{-0.1in}
\label{fig:visualization-dataset}
\end{figure*}


\begin{table}[!tbp]
\centering
 \caption{Comparison of our something-STAR dataset with existing spatio-temporal activity detection datasets. }
\scalebox{0.78}{
\begin{tabular}{c|c|c|c|c|c|c}
\hline
 &
   \begin{tabular}[c]{@{}c@{}}Something-\\ STAR\end{tabular} &
  UCF101-24~\cite{soomro2012ucf101} &
  DALY~\cite{weinzaepfel2016human} &
  AVA~\cite{gu2018ava} &
  J-HMDB~\cite{jhuang2013towards} &
  UCF-Sports~\cite{rodriguez2008action} \\ \hline
\#classes                 & 47          & 24      & 10       & 80      & 21       & 10      \\
action types              & hand-object & sports  & everyday & movie   & everyday & sports  \\
\#clips                   & 14100       & 3207    & 8133     & 430     & 928      & 150     \\
avg resolution            & 396x240     & 320x240 & 1290x790 & 640x480 & 320x240  & 690x450 \\
avg video dur.            & 3.6s        & 5.8s    & 3min 45s & 15min   & 1.4s     & 5.8s    \\
avg action dur.           & 2s          & 4.5s    & 7.9s     & 15min   & 1.4s     & 5.8s    \\
action duration ratio     & 55.6\%      & 78\%    & 4\%      & 100\%   & 100\%    & 100\%   \\
average \#instances/class & 300         & 168     & 364      & 5       & 44       & 15      \\
spatial annotation        & all         & all     & sparse   & all     & all      & all     \\ \hline
\end{tabular}

}
\vspace{-0.1in}
\label{tab:comparision_between_datasets}
\end{table}

\begin{figure*}[]
    \centering
    \includegraphics[width=0.45\linewidth]{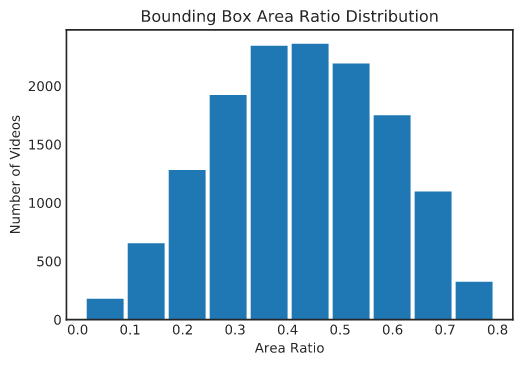}
    \includegraphics[width=0.45\linewidth]{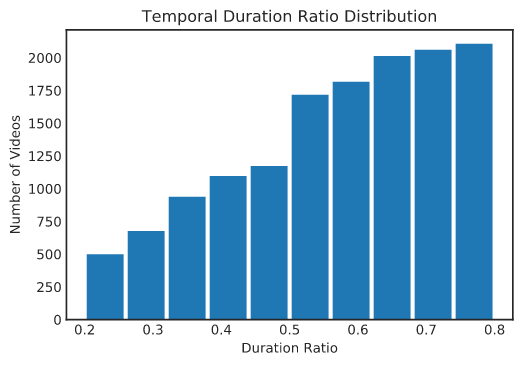}
    \vspace{-0.1in}
    \caption{Distribution for average bounding box area ratio and temporal duration ratio in our something-STAR dataset. }
    \label{fig:dataset-stats}
    \vspace{-0.1in}
\end{figure*}

To get the action tube annotation suitable for the spatio-temporal action detection, we start from the raw object bounding box annotation of ``something-something" dataset~\cite{materzynska2019something}.
We first take the union of relevant object bounding boxes involved in the action to form action tubes.
Then we calculate two thresholds for the action tubes to select videos used in our something-STAR action detection dataset.
One temporal threshold is the ratio of the action duration to the whole video duration, and the other spatial threshold is the average ratio of the action box area to each frame area.
We only keep videos with temporal threshold in the range of 0.2-0.8  and spatial threshold in the range of 0.01-0.8.
We also balance the number of videos in each class and set the number of videos in each class to be 300. We discard the classes with videos less than 300, and sample 300 videos per class for classes with more than 300 videos.
This results in a total of 47 classes and 14100 videos in our proposed ``something-STAR" dataset.
Note that there also exists some other video datasets with raw object bounding box annotation, e.g. EPIC-Kitchens dataset~\cite{damen2018scaling}.
However EPIC-Kitchens dataset uses head-mounted camera with heavy camera motion which might disturb action motion learning.
In this paper, we take ``something-something" dataset with static camera and focus on designing models to learn the action motion pattern for saptio-temporal action detection.


We also visualize one example video from each spatio-temporal action detection dataset (AVA, DALY, UCF101-24 and our something-STAR dataset) in Figure~\ref{fig:visualization-dataset}.
From the first row of Figure~\ref{fig:visualization-dataset}, you can see that the action temporal duration in AVA dataset~\cite{gu2018ava} lasts from the video beginning to the end.
Another dataset -  DALY dataset~\cite{weinzaepfel2016human} shown in the second row of Figure~\ref{fig:visualization-dataset} annotates the action temporal duration but within the action temporal duration they only selectively annotate one to five frames.
UCF101-24~\cite{soomro2012ucf101} shown in the third row of Figure~\ref{fig:visualization-dataset} is the only relative complete dataset with both spatial and temporal annotation in the form of action tubes.
In our proposed something-STAR dataset, we focus on modeling the action with multi-object interaction and try to capture the motion pattern for spaito-temporal action detection.
Notably the annotated action in DALY dataset also contains the human and object interaction, but since they choose human as action subject, the involved object (e.g. cup) is occluded by the human body, resulting in the spatial localization heavily relying on the human detector.
In our dataset, we have hand as the action subject and alleviate the spatial occlusion problem in subject-object interaction modeling to a certain extent.

In Table~\ref{tab:comparision_between_datasets}, we list the summary information for our proposed ``something-STAR" dataset as well as several other action detection datasets.
Our something-STAR dataset contains 14100 videos across 47 action labels with balanced 300 videos per class. 
Figure~\ref{fig:dataset-stats} shows the distribution of tube temporal duration ratio and the average spatial box area ratio in our something-STAR dataset.
While for half of the classes in the UCF101-24 dataset, the action lasts for more than 80\% of the video duration, the action tubes in our something-STAR dataset have reasonable temporal duration ratio distribution for testing the temporal localization capability of the spatio-temporal action detection models. The average temporal duration ratio of the action tubes in our something-STAR dataset is 55.6\%.
Moreover, compared to UCF101-24 datset with 24 action classes, our dataset is on a larger scale with 47 action classes.
The list of classes in our proposed something-STAR dataset are shown in the supplementary material.

\vspace{-5pt}
\section{Experiments}
We report results on our proposed spatio-temporal action detection dataset- something-STAR with multi-object interaction, using our STAR model with simultaneous spatial and temporal regressions.
We also evaluate our STAR model on the standard person centric, spatio-temporal action detection benchmark UCF101-24~\cite{soomro2012ucf101} and show the generalization of our model with very competitive results.
Sections~\ref{exp:something-STAR_res} and~\ref{exp:UCF101-24_res} provide the experimental details and evaluation results on these two datasets, and section~\ref{exp:speed_res} provides the detection speed comparison.

The spatio-temporal action detection results are evaluated in terms of video mean Average Precision - video mAP@$\alpha$ where $\alpha$ denotes different Intersection over Union (IoU) thresholds for action tubes, as is the common practice in the literature.
The IoU for the predicted and ground truth action tubes is defined as the product of the temporal IoU between the time segments of the two action tubes and the average spatial IoU on the frames where both action tubes are present. 
The candidate action tube detection is correct if its intersection with the ground action tube is above a threshold (e.g. $0.2$ or $0.5$) and the action class is predicted correctly.
We also report the frame mAP@$\alpha=0.5$ focusing on evaluating the bounding box spatial localization in action tube, where $\alpha$ denotes bounding box IoU threshold.
Additionally, we evaluate the action tube from the temporal localization aspect
following the evaluation metric used in the temporal activity detection task, namely average mAP at temporal IoU thresholds $\alpha \in (0.5,0.95)$ with step 0.05, mAP@$\alpha=0.5$ and mAP@$\alpha=0.7$.

\begin{table}[!t]
\centering
\caption{Detection results on Something-STAR dataset in terms of video mAP@0.2, video mAP@0.5, frame mAP@0.5, temporal mAP@=0.5, temporal mAP@0.7 and average temporal mAP at ten evenly distributed thresholds between [0.5:0.95] (in percentage). }
\begin{tabular}{l || c c | c |c c  c } 
\hline
 ~ & \multicolumn{2}{c|}{video mAP} & \multicolumn{1}{c|}{frame mAP} &  \multicolumn{3}{c}{temporal mAP} \\ 
 ~ & 0.2  & 0.5  & 0.5  & 0.5 & 0.7  & [0.5:0.95]  \\ \hline
\modelname - sport1M & 44.72 & 16.65 & 23.53 & 50.30 & 37.38 & 29.25  \\  
\modelname - something-finetuned & 51.00 & 22.42 & 29.22 & 57.44 & 44.33  & 34.65  \\ \hline  
\end{tabular}
\vspace{-0.1in}
\label{res:activitynet}
\end{table}


\subsection{Something-STAR Dataset}
\label{exp:something-STAR_res}
Our proposed something-STAR dataset for action detection with multi-object interaction consists of 47 action classes of 14100 videos with 80\% for training and 20\% for testing.
We take 5\% videos from the train set as held out validation to get the best hyperparameter setting, and re-train the best model on the whole train set and report final results on the test set. 
Our model accepts an input video buffer of 80 frames at 12 frames per second (fps).
Data augmentation strategies except horizontal flip are not applied during training.

We run the first baseline of our STAR model starting from Sports-1M pretrained C3D classification weights released by the author in~\cite{tran2015learning} and get the video mAP@0.5 of 16.65\%.
We also finetune C3D model on the something-something classification data and get 25.6\% top-1 classification accuracy and 50.6\% top-5 classification accuracy.
Starting from the something-something finetuned C3D classification backbone, our STAR model's video mAP@0.5 result further improves by 5.77 points and achieving 22.42\%. 
This indicates that our model can benefit from better pretrained video classification backbone during end-to-end training.

\begin{figure*}[!t]
\centering
\includegraphics[width=0.9\linewidth]{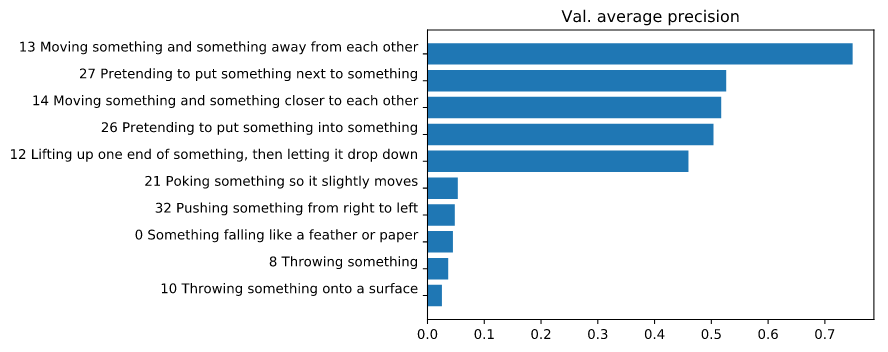}
\vspace{-0.1in}
\caption{Visualization of the top 5 action classes with highest average precision and the top 5 action classes with lowest average precision from our STAR model on our something-STAR dataset.}
\vspace{-0.1in}
\label{fig:visualization-ap}
\end{figure*}

In this paper, we study the spatio-temporal action detection with multi-object interaction, and we choose hand as subject as opposed to human to alleviate the object occlusion problem to certain extent.
However, we also notice that the appear of hand has a high correlation with the action temporal duration. To investigate the effect of hand prior for action temporal localization,
we evaluate the temporal proposals localized by the sate-of-the-art hand detector~\cite{Shan2020Hand} in class agnostic way and achieve 73.81\% using the AR/AN metric between IoU thresholds 0.5 and 0.95 as is the standard evaluation paradigm for the temporal localization task in ActivityNet challenge.
Our STAR model starting from sport1M classification weights achieves 75.37\% using the same metric.
This indicates that even though the hand has high correlation with the action temporal segment, our STAR model relying on direct temporal regression still outperforms in temporal localization.

Figure~\ref{fig:visualization-ap} shows the top 5 action classes with highest average precision and the top 5 action classes with lowest average precision on this dataset.
It seems that our STAR model generally outperforms in actions with two objects interacting with each other.
Figure~\ref{fig:something-good-example} and \ref{fig:something-bad-example} show two representative qualitative results from two videos.

\vspace{-5pt}
\subsection{UCF101-24 Dataset}
\label{exp:UCF101-24_res}

\begin{table}[!t]
\centering
\caption{
Frame rate and spatial resolution ablation results for our STAR model on UCF101-24 dataset in terms of video mAP@0.5 (\%). }
\medskip
\begin{tabular}{ l l || c }  
\hline
 temporal frame rate & spatial resolution &  video mAP@0.5 \\ \hline
25 & 176$\times$ 176  & 50.64  \\ 
50 & 176$\times$ 176  &  43.51  \\ 
12.5 & 176$\times$ 176  & 40.59 \\ 
25 & 112$\times$ 112  & 40.89 \\ 
25 & 224$\times$ 224  & 44.22 \\ \hline
 Full model (ensemble) & & 53.0  \\ \hline
\end{tabular}
\vspace{-0.11in}
\label{table:Ablation:THUMOS14}
\end{table}

The spatio-temporal action detection dataset UCF101-24 is a subset of 24 action classes from the original UCF101 dataset~\cite{soomro2012ucf101} for action classification.
We use the recently revised ground truth action tube annotation~\cite{singh2017online}.
It consists of 3207 videos with 2293 for training and 914 for testing.
Our model accepts an input video buffer of 384 frames.
Data augmentation strategies are not applied during training except horizontal flip.
We initialize the 3D ConvNet part of our model with C3D weights trained on Sports-1M and finetuned on UCF101 released by the author in \cite{tran2015learning}.

\begin{figure*}[!t]
\centering
\subfigure[Something-STAR good example: Moving something and something closer to each other]{
\hbox{\hspace{0.25cm}\includegraphics[width=10.52cm]{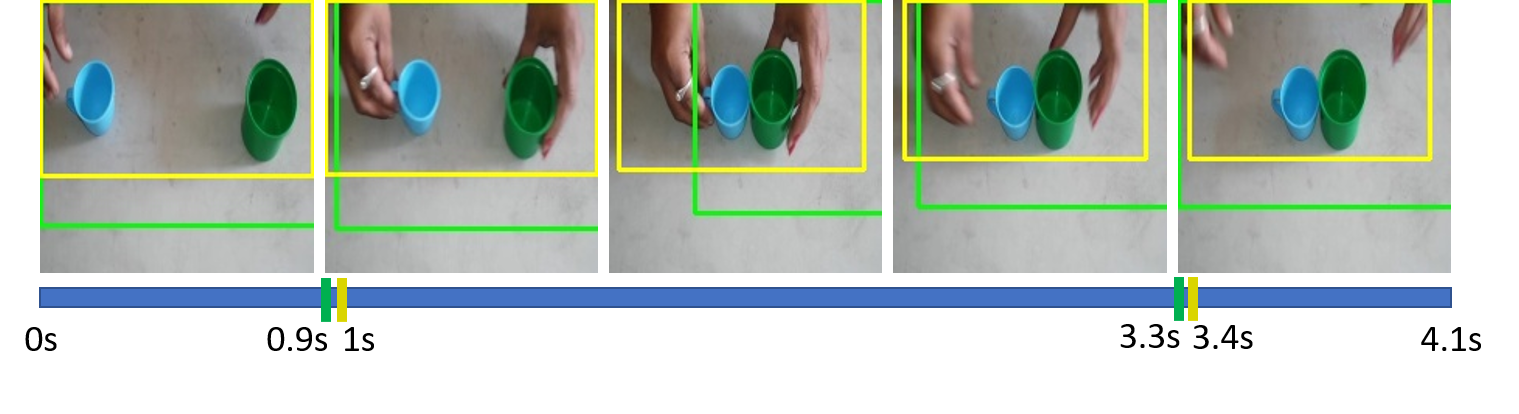}}
\label{fig:something-good-example}
}
\subfigure[Something-STAR bad example: Pushing something with something]{
\hbox{\hspace{0.1cm}\includegraphics[width=0.85\linewidth]{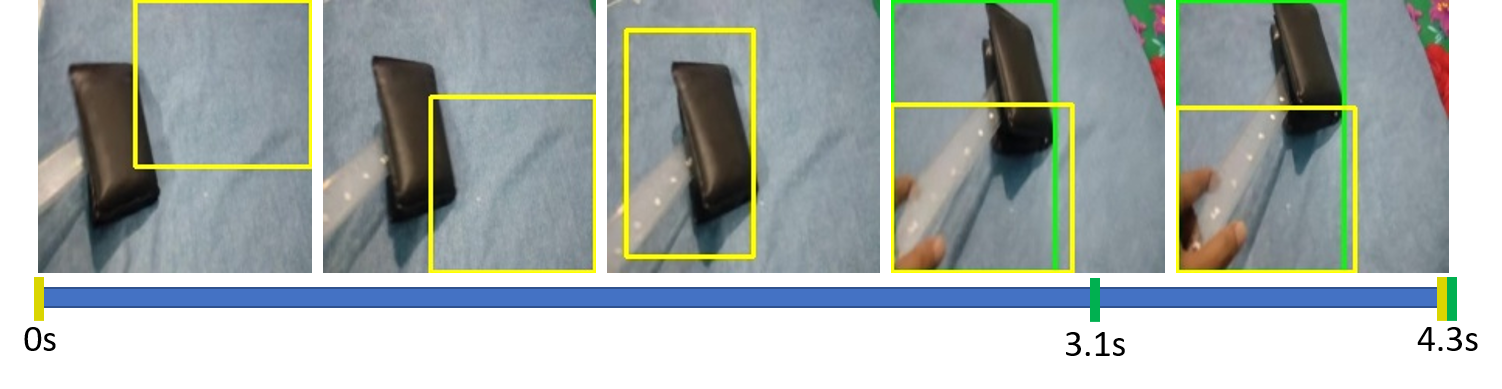}}
\label{fig:something-bad-example}
}
\subfigure[UCF101-24 good example: Cricket Bowling]{
\includegraphics[width=0.85\linewidth]{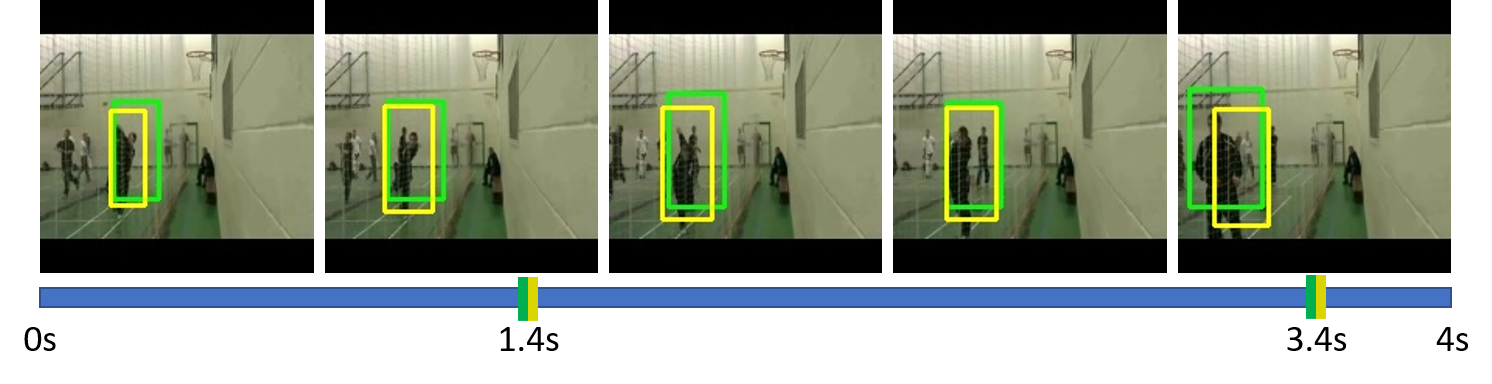}
\label{fig:ucf-good-example}
}
\subfigure[UCF101-24 bad example: Salsa Spin]{
\hbox{\hspace{0.3cm}\includegraphics[width=0.86\linewidth]{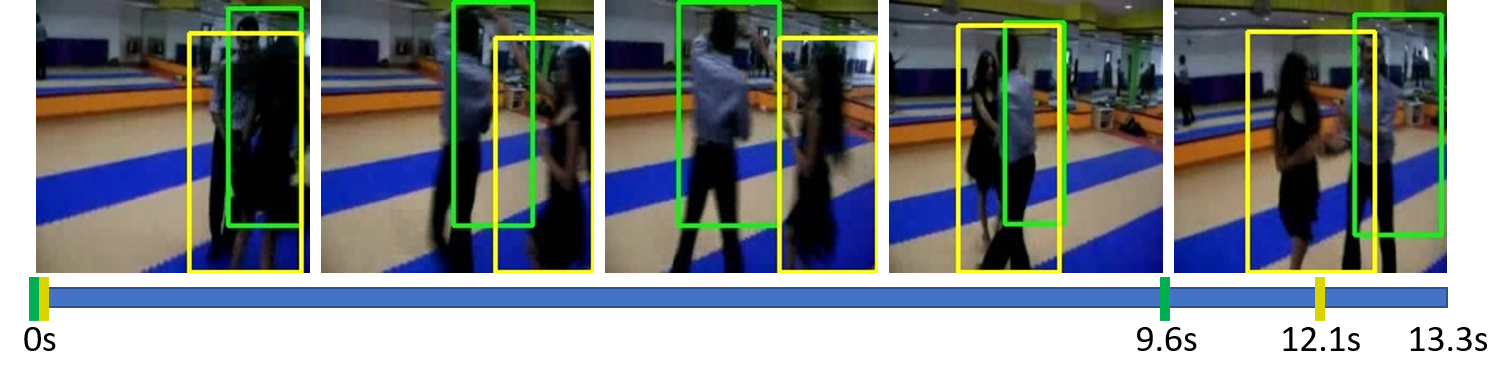}}
\label{fig:ucf-bad-example}
}
\vspace{-0.1in}
\caption{Qualitative visualization of our STAR model detection on our something-STAR dataset and UCF101-24 dataset - green is ground truth, yellow is prediction.}
\vspace{-0.1in}
\label{fig:visualization-samples}
\end{figure*}

Since our model is built on top of the shared three dimensional fully convolution features,
the  spatial and temporal resolution can be increased or shrinked correspondingly.
We explore different combinations in Table~\ref{table:Ablation:THUMOS14} and find the best configuration of 25 fps and $176\times176$ spatial resolution.
This single model achieves the video mAP@0.5 of 50.64\% and the ensemble of all the five models gives us the video mAP@0.5 of 53.0\%.
This result is very competitive when comparing to other sate-of-the-art spatio-temporal action detection models in Table~\ref{tab:other-works} considering that most of the models in this table use more advanced two-stream video classification backbones~\cite{simonyan2014two,carreira2017quo}.
Our model could potentially benefit from a more advanced video classification backbone (as shown in previous section~\ref{exp:something-STAR_res} that STAR model trained from something-something fine-tuned classification backbone got improved results compared to sport1M classification weights). 
Our model doesn't have a very good frame mAP, but our model's spatial and temporal evaluation metric, video mAP@0.5, is still high showing the superiority of our model's temporal localization ability.
Notably, the action tube annotation in the UCF101-24 benchmark is mostly person centric. 
Though the spatial localization module in our STAR model is designed to capture the action motion among multiple objects by regressing one big action bounding box for enclosing multi-object interaction, our STAR model could be easily deployed on single object/person centric action detection dataset (e.g. UCF101-24) and achieve competitive results compared with the previous models relying on off-the-shelf object/person detectors, showing the generalbility of our proposed STAR model with simultaneous spatial and temporal regression for jointly action detection.
Figure~\ref{fig:ucf-good-example} and \ref{fig:ucf-bad-example} show two representative qualitative results from two videos. For the class ``Salsa Spin", our STAR model focuses on localizing the area with two persons but the GT box is annotated with single person.

\begin{table}[!t]
\centering
\caption{ Spatio-temporal action detection results on UCF101-24 dataset evaluated in video mAP@0.2, video mAP@0.5 and frame mAP@0.5 (in percentage). }
\begin{tabular}{l||cc|c}
\hline
\multirow{2}{*}{}                                                & \multicolumn{2}{c|}{video mAP} & frame mAP \\
                                                                 & 0.2            & 0.5           & 0.5       \\ \hline 
Peng, et al~\cite{peng2016Multi-region}                                                     & 42.3         &  -           & 39.6       \\ 
Hou, et al~\cite{hou2017tube}                                                       & 47.1           & -             & 41.4      \\
\begin{tabular}[c]{@{}l@{}}Weinzaepfel, et al~\cite{weinzaepfel2016human}\end{tabular} & 58.9           & -             & -         \\
STEP~\cite{yang2019step}                                                             & 76.6           & -             & 75.0      \\
Saha, et al~\cite{saha2016deep}                                                      & 66.8           & 35.9          & -         \\
Gurkirt, et al~\cite{singh2017online}                                                     & 73.5           & 46.3          & -         \\
Pramono, et al~\cite{pramono2019hierarchical}                                                   & 80.4          & 49.5          & 73.7     \\
\begin{tabular}[c]{@{}l@{}}Chéron, et al~\cite{cheron2018flexible}\end{tabular}         & 76.0           & 50.1          & -         \\
Zhao, et al~\cite{zhao2019dance}                                                      & 78.5          & 50.3         & -         \\

ACT~\cite{kalogeiton2017action}                                                              & 77.2           & 51.4          & 67.1         \\
TACNet~\cite{song2019tacnet}                                                           & 77.5           & 52.9          & 72.1      \\
Gu, et al~\cite{gu2018ava}                                                              & -              & 59.9          & 76.3         \\ \hline 
STAR model (ours) & 77.9 & 53.0 & 63.0 \\ \hline 
\end{tabular}
\vspace{-0.1in}
\label{tab:other-works}
\end{table}

\vspace{-2pt}
\subsection{Spatio-temporal Activity Detection Inference Speed}
\label{exp:speed_res}
The inference speed for our STAR model is very fast due to the shared feature encoding and the simple action tube construction policy during testing without heavy post-processing.
Our model can make test inference at 250 fps including the action tube construction on a single TITAN X (Pascal) for spatial resolution $176\times176$, which is a significant improvement compared to previous models~\cite{pramono2019hierarchical,kalogeiton2017action,singh2017online} with inference speed under 40 fps.

\vspace{-2pt}
\section{Conclusion}
In this paper, we explore the problem of spatio-temporal action in videos with multi-object interaction. We propose a simple and effective spatio-temporal action detection model called STAR with both spatial and temporal regression during training. Our model directly connects the boxes within the predicted temporal duration during inference without heavy post-processing and achieves fast detection speed.
Additionally, we propose a new large-scale spatio-temporal action detection dataset with multi-object interaction in the annotated action tube for public use.
We validate the effectiveness of our proposed STAR model on both our proposed dataset and the UCF101-24 benchmark, and we expect more future work along this direction.


%
%
\bibliographystyle{splncs04}
\bibliography{egbib}
\end{document}